\documentclass[10pt, fleqn]{article}

\usepackage{graphicx}
\usepackage{siunitx}
\usepackage{longtable,tabularx}
\usepackage{comment}
\usepackage{subfig}
\usepackage{verbatim}
\usepackage{float}
\usepackage[utf8]{inputenc}
\DeclareUnicodeCharacter{2212}{$-$}
\usepackage{graphicx}
\usepackage[fleqn]{amsmath}
\usepackage[version=4]{mhchem}
\usepackage{siunitx}
\usepackage{longtable,tabularx}
\usepackage{nccmath}
\usepackage{amsmath}
\usepackage[superscript]{cite}
\usepackage[margin=1in]{geometry}
\usepackage[toc,page]{appendix}
\usepackage{setspace}
\usepackage{amssymb}
\usepackage{tabu}
\usepackage{listings}
\usepackage{afterpage}
\usepackage{titlesec}
\usepackage[labelfont=bf]{caption}
\usepackage{fancyhdr}
\usepackage{pgf}
\usepackage{float}
\usepackage{cleveref}
\usepackage{textcomp}
\usepackage{pict2e}
\usepackage{wasysym}
\usepackage{multicol}
\usepackage{lipsum}
\usepackage{indentfirst}
\newenvironment{Figure}
  {\par\medskip\noindent\minipage{\linewidth}}
  {\endminipage\par\medskip}

\usepackage[utf8]{inputenc}

\titleformat{\section}
{\normalfont\normalsize\bfseries}{\thesection}{0.5em}{}
\titleformat{\subsection}
{\normalfont\normalsize\bfseries\em}{\thesubsection}{0.5em}{}
\titleformat{\subsubsection}
{\normalfont\normalsize\bfseries\em}{\thesubsubsection}{0.5em}{}
\begin{document}

\begin{flushright}
\Large 

\textbf{[SSC23-VII-02]}
\end{flushright}
\begin{centering}      
\Large 

\textbf{Development of On-Ground Hardware In Loop Simulation Facility for Space Robotics }\\
\vspace{0.5cm}
\normalsize 

{\textbf{Roshan Sah}}, \textbf{Raunak Srivastava}, \textbf{Kaushik Das}\\
{Space Systems, Tata Consultancy Services(TCS)-Research}\\
{Bangalore, India}\\
{sah.roshan@tcs.com}

\vspace{0.5cm}
\centerline{\textbf{ABSTRACT}}
\vspace{0.3cm}
\end{centering}

Over a couple of decades, space junk has increased rapidly, which has caused significant threats to the LEO operation satellites. A mitigating measure should be taken to protect the LEO space environment. An Active Debris Removal (ADR) concept continuously evolves for space junk removal. One of the ADR methods is Space Robotics, whose function is to chase, capture and de-orbit the space junk. This paper presents the development of an on-ground space robotics facility in the TCS Research for on-orbit refueling and debris capture experiments. A Hardware-in-Loop Simulation (HILS) system will be used for integrated system development, testing, and demonstration. HILS is the most effective and vital system to test the on-orbit docking mechanism's reliability, usability, and safety. The HiLS test facility of TCS Research Lab will use two Universal Robot(UR)5e and UR10 manipulators in which one manipulator is attached to the robotic-RG2 gripper, and the other is attached to a force-torque sensor named Hexa-E Onrobot and with a scaled mock-up satellite model. The first UR5 manipulator will be mounted on a 7-axis linear rail and contain the docking probe. First UR5 manipulator with the suitable gripper has to interface its control boxes. The grasping algorithm was run through the ROS interface line to demonstrate and validate the On-orbit and Debris removal operation. The manipulator will be mounted with LIDAR and a Real sense camera to visualize the mock-up model, find the target model's pose and rotational velocity estimation, and a gripper that will move relative to the target model. The other manipulator has the UR10 control, providing rotational and random motion to the mock-up satellite, enabling a dynamic simulator fed by force-torque data. The dynamic simulator is fed up with the orbit propagator model SGP4, which will provide the orbiting environment to the target model. For the simulation of the docking and grasping of the target model, a 7-axis linear rail of a 6-meter setup is still in the procurement process. Once reaching proximity, the grasping algorithm will be launched to capture the target model after reading the random motion of the mock-up satellite model. The HILS system proposed in this paper helps develop on-orbit servicing (OOS) like repairing, upgrading, transporting, rescuing technologies, on-orbit refueling, and berthing and debris removals. 
   
\begin{multicols*}{2}

\section*{INTRODUCTION}
The growth of spacecraft launches has increased substantially over the past couple of decades, and the statistics results indicated an average of more than one hundred satellites were launched annually. Most spacecraft launches have accomplished their mission goals, but some experienced failures and irregularities \cite{david_2001}. In the past, a launcher failure is the most typical failure of the spacecraft. However, an on-orbit failure has recently overtaken launch failures, resulting in billions of dollars lost in space organizations like NASA, ESA, ISRO, etc. Eventually, all launched satellites run out of fuel after their end-of-life (EOL) span, which causes the formation of space junks/debris at different orbit regions.

In recent decades, the spaces junks (residual of the spent satellite, rocket staging, body and booster, junk particles from the collision of debris objects) have increased drastically in Low Earth Orbit (LEO) region, which can cause a severe threat to the operation satellite at LEO orbit. Several works of literature have shown that servicing, like repairing, refueling, upgrading, etc., of damaged spacecraft in flight is cost-effective and extends the existing satellite's life span. In addition, the post-mission removal of the spent satellite, rockets staging, etc.,  will keep the operational satellite in a favorable orbital environment for upcoming space missions. An On-Orbit Servicing (OOS) has become instrumental in increasing the life span of many satellites, thereby helping in the sustainable utilization of space orbits.

To do the servicing of damaged spacecraft, NASA realized, as early as the 1980s, the significance of the robotics on-orbit servicing activities to protect assets in space \cite{ELLERY2008632}. It was termed Space Robotics, which is the combination of spacecraft and robotics manipulators. Recently, Roshan et al. have developed a spacecraft with single/dual manipulators, which are termed a “Debris Chaser Satellite” \cite{AIAA_2022_Roshan}\cite{AIAA_2022_Raunak} \cite{ASET2022} whose primary function is to chase, capture, and de-orbit\cite{AIAA_2023_Roshan} the space junks to a shallow earth orbit. However, the debris chaser satellite can also be used for servicing applications. Autonomous control of space robots for precise in-orbit operations (satellite servicing, active debris removal, etc.) has become critical for long-term sustainable use of the Earth’s orbits due to a multi-fold increase in space debris. However, the spacecraft-manipulator couple system produces non-linearity to space systems’ complex control problem. Manipulator motion induces reaction forces and moments on the satellite, which disturbs its position and attitude, affecting the end-effector’s pose. Therefore, the dynamic interaction between the manipulator and the base satellite must be considered while designing a control law for such space manipulators. 

The space robot consists of three different modes of approach\cite{IEEE_2022};

1. Free Flying space robot: 

It refers to the class of satellite-manipulator systems wherein the cold gas thrusters/jets facilitate\cite{SSC2022_1} \cite{SSC2022_2} a stabilized and controlled base for robotic manipulator motion. The fixed base is favorable for trajectory formation of manipulators, however, at the expense of fuel consumption and restricted workspace.

2. Free Floating space robot:

It refers to the class of satellite-manipulator systems wherein the combined system momentum remains conserved due to the absence of an external force or torque in the terminal phase when the arm is deployed to reach the target. Only the manipulator arm motion is actively controlled in a Free Floating configuration. At the same time, the satellite base is left uncontrolled to freely translate and rotate as per the forces and moment acted upon by the arms motion. In contrast, there is no fuel consumption at all.

3. Rotation Floating Space Robot: 

Rotation floating space robots\cite{IEEE_2023} are a comparatively newer approach wherein the base satellite attitude is controlled using actuators like reaction wheels (RW) which preserve the system’s momentum and control the manipulator’s arms. The satellite base is still left free to translate. This approach causes fuel consumption in a limited amount.

These three approaches are primarily used for the manipulation in orbit for OOS applications. Ground-based experiments will be required to verify and explore the capability, feasibility, and limitations of the critical trajectory planning and control algorithm before servicing the satellite i.e., the space robot is launched for OOS missions. A ground-based experiment helps us to ensure that the space robot can successfully implement the mission tasks like orbit capturing and docking operations. Hence, setting up a suitable ground-based experiment system for a space robot performing fine manipulation becomes essential.

The ground-based experiment for the evaluation of planning and control algorithms for space robots is challenging in two ways; 

1. the  coupled space robot system needs to be permitted to move in the 6DOF system,  and

2. the effect of gravity cannot be canceled as space considers the micro-gravity environment.

Over the last decade, there have been five methods to emulate the micro-gravity environment on the ground \cite{ground_expt_2012};

1. Free fall motion or airplane flying at the stratosphere.

2. Neutral buoyancy

3. suspension system

4. air-bearing table.

5. Hybrid method.

The first four methods to emulate the micro-gravity environment are expensive and time-consuming. Therefore most of the space research organizing focuses on the hybrid method, which is the mathematical model combined with the mechanical model. The trajectory of the space robot in a micro-gravity environment is calculated using the gravity-based mathematical model. Then the mechanical model is forced to move according to the calculation. A precise dynamic model can emulate the behaviors of a space robot coupled system. The hybrid method is relatively cheap and can be set up using ground-based robots like UR5, UR10, and KUKA.

Organizations like NASA, ESA, DLR (German Aerospace Center), Surrey Space Center, and JAXA have set up several research facilities to execute the ground-based experiment to verify the space robotics operation for OOS applications. The DLR Robotics Laboratory has set up a European Proximity Operations Simulator (EPOS),  which consist of two KUKA robot in which; one is used for simulating target and the other for simulation chaser, linear rail, sun simulator, and sensors. The EPOS facility can be used to test the autonomous GNC for rendezvous and mission inspection\cite{benninghoff2017european} \cite{aerospace8090235}. Similarly, the STAR lab of the University of Surrey has developed the Orbital robotics testbed, which is used to simulate the 6DOF manipulation by using multiple robotics arms or air-bearing tables. Their setup has also considered orbital disturbance and tries to grasp the 3D printed mock-up model using a range of control modes from tele-operation to full autonomy \cite{On-OrbitRobotic_Nikos}.

In this paper, we have established on ground Hardware In Loop Simulation (HiLS) facility to explore and verify the capability of the space robot to execute fine manipulation, non-cooperative target capturing,  and the end phase of docking. This HiLS facility is the first space robotics laboratory in India set up by TCS Research India and will be used for future ISRO  OOS missions. Using ground-based robots, a hybrid simulation approach has been used to verify our trajectory and control algorithms. These verified algorithms can be used for upcoming missions.

The rest of the paper has been organized into the following sections; Section: Hybrid Simulation Method describes the simulation and hardware concept of dynamics emulation of chaser-target motion using the Pybullet and URs manipulator. The section: Space Robotics Laboratory(SRL) explains about objective, lab layout, and the different SRL components required for on-ground experiment systems. In the section: Results and Discussion, the initial results of the Hardware-in-loop System (HiLS)of URs replicating the chaser's and target's motion were explained. The last section deals with the conclusion and future works that will be executed in the coming days by considering the multiple scenarios of space rendezvous and docking in a more real-time consideration.


\section*{HYBRID SIMULATION METHOD}
Figure \ref{Fig:1} shows a snapshot of the PyBullet environment created to simulate the chaser-target space robot. A red target is being tracked by a Rotation Floating space robot\cite{SRIVASTAVA2022147} with a UR5 manipulator on-board it. Three orthogonal reaction wheels are mounted on the space robot to control the attitude of the base satellite. The objective is to use a two-layered MPC-based control \cite{AIAA_2023_Raunak} architecture that controls the motion of the satellite base and UR5 manipulator to enable it to reach the moving target. After that, the target will be grabbed using an impedance controller, followed by bringing the arm (with the target) back to the initial configuration. PyBullet simulations have been performed, and the results are shown later. An orbital environment \cite{SSC_2021_Roshan} \cite{SSC_2021_Raunak} needs to be incorporated with the Pybullet environment by using STK tools which will be part of the future work.

\begin{Figure}
    \centering
   \includegraphics[width = 0.97\linewidth, height = 2.7 in]{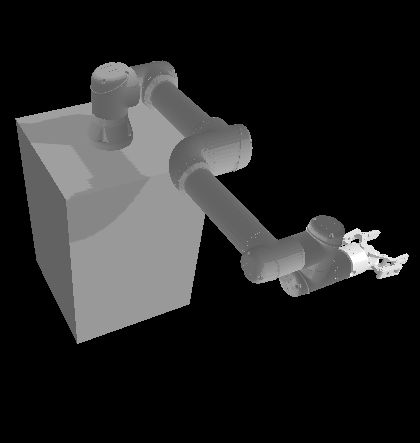}
    \captionof{figure}{Space Robot in Pybullet Environment.}
    \label{Fig:1}
\end{Figure}

The dynamic equation of the space robot is generally expressed in the following form;
\begin{equation}
   \begin{bmatrix}
     H_b & H_{bm} \\
      H_{bm}^T & H_m 
    \end{bmatrix}
    \begin{bmatrix}
     \Ddot{x_b}   \\
      \Ddot{\Theta_s}  
    \end{bmatrix}   
    + \begin{bmatrix}
     c_b   \\
      c_m  
    \end{bmatrix} 
    = \begin{bmatrix}
     F_b   \\
      \tau_m  
    \end{bmatrix} 
    +\begin{bmatrix}
     J_b^T   \\
      J_m^T  
    \end{bmatrix}   
    \label{eqn. 1}
\end{equation}

whereas

$H_b \in R^{6x6}$ and $H_m \in R^{nxn}$ are the inertial matrices of the base and the manipulator's arm; 

$H_bm \in R^{6xn}$ is the couple inertial matrix;

$c_b \in R^6$ and $c_m \in R^6$ are the velocity-dependent non-linear terms for base and arm;

$F_b, F_e, \tau_m \in R^{6}$ are force and torques on the centroid of the base and exert on the manipulator's hand, torque on the manipulator's joints.

\begin{equation}
  \centering
  \Dot{x_e} = J_b \Dot{x_b} + J_m \Dot{\Theta_s}
   \centering
  \label{eqn. 2}
\end{equation}
Whereas,
$J_b$ and $J_m$ are the Jacobian matrices dependent on the motion of the base and the manipulator;
$\Dot{x_b}$ and$\Dot{x_e}$ are the absolute velocity of the space base and end-effectors.
$\Theta_s$ is the joint angle of the space manipulator.

Since no external forces and torques are acting on the free-floating system of the space robot, the linear and angular momentum is considered zero initially. Therefore, the system momentum keeps zero according to the conservation law. The equation of such a free-floating space robot is given as

\begin{equation}
  \centering
  H_b\Dot{x_b} + H_bm \Dot{\Theta_s} = 0
   \centering
  \label{eqn. 3}
\end{equation}

Hence, the absolute velocity of the space robot can be solved from equation \ref{eqn. 3} as

\begin{equation}
  \centering
  \Dot{x_b} =-H_b^{-1} H_bm \Dot{\Theta_s} = H \Dot{\Theta_s}
  \centering
  \label{eqn. 4}
\end{equation}
Substituting the value of the absolute velocity of space base for equation \ref{eqn. 4} to \ref{eqn. 3}, the following relationship can be formed,

\begin{equation}
  \centering
  \Dot{x_e} =(J_m-Jb H_b^{-1} H_bm )\Dot{\Theta_s} = J^* (\Phi_b , \Theta_s)\Dot{\Theta_s}
  \centering
  \label{eqn.5}
\end{equation}

Whereas,
 $J^* (\Phi_b , \Theta_s)$ is the Generalized Jacobian Matrix (GJM)

 And by GJM, the manipulator's hand can be controlled by resolved motion rate or acceleration control in the inertial space. 

The simulations are followed by hardware experiments to validate the results obtained from PyBullet. Towards this end, a Hardware-in-the-loop-based Hybrid approach is followed, the most commonly used approach for space robots. Figure \ref{Fig:2} shows the philosophy behind the hardware experiments. The motion of the chaser end-effector and the target is calculated concerning a frame attached to the base of the chaser satellite. The transformed 6 DOF motion is replicated on two Universal Robot arms: Chaser on a UR5 arm and target motion on the UR10 arm. The actual motion of the hardware arms is then fed back to the PyBullet environment as real-time feedback. The 6 DOF motion is transformed back into the inertial frame, which is then acted upon by the proposed controller. The cycle repeats until and unless the chaser arm (UR5) reaches the target arm (UR10). In the preliminary results, independent motions of the UR5 (chaser motion) and UR10 (target motion) were executed in sync with the PyBullet environment in real time. Both arms produced perfect results and reproduced the PyBullet results. In the future, they will be run together to complete the pipeline executed in PyBullet.

\begin{Figure}
    \centering
   \includegraphics[width = 0.99\linewidth, height = 2.0 in]{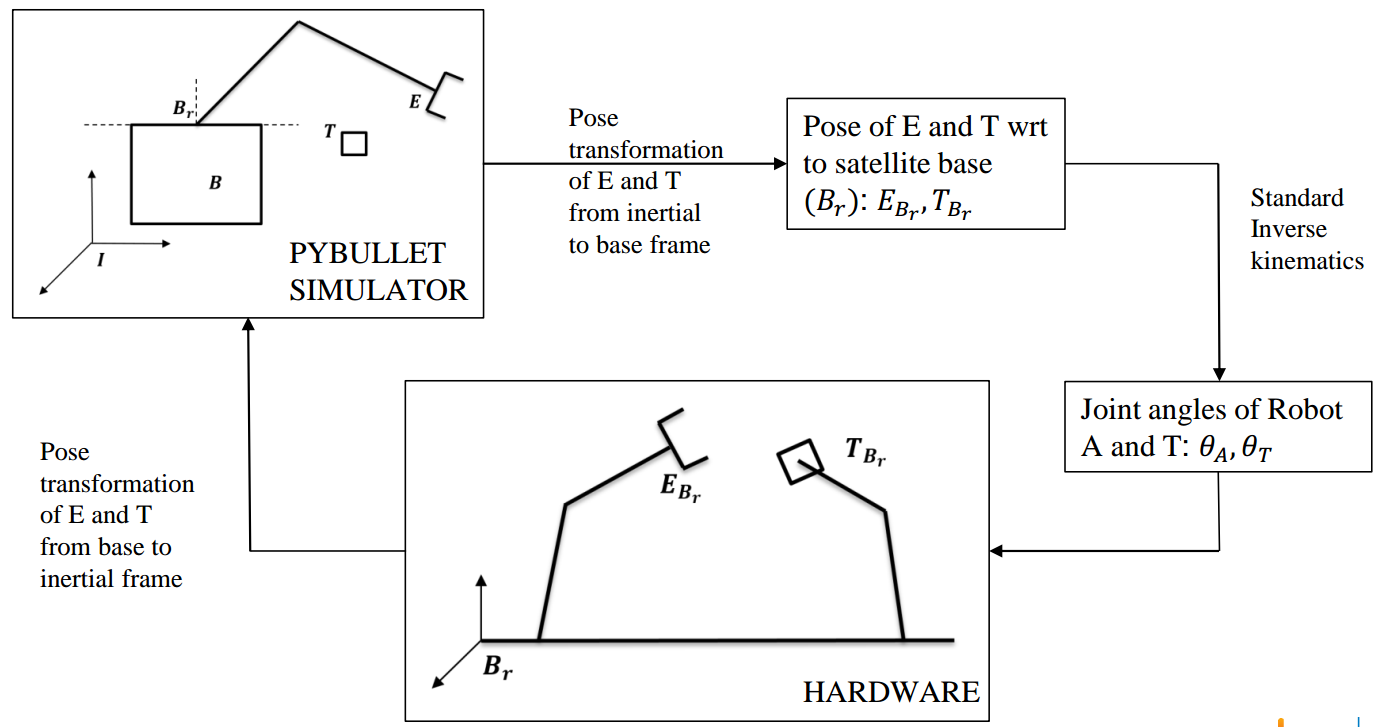}
    \captionof{figure}{Hybrid Approach of Hardware in the loop.}
    \label{Fig:2}
\end{Figure}


\section*{SPACE ROBOTICS LAB(SRL) SETUP}

The space robotics lab is developed under a cross-section area of 12.2m x 5.8m in the TCS Research Building. The concept of a space robotics lab was taken from the German Aerospace Center(DLR) \cite{roa2017robotic} robotics test facility where they developed European Proximity Operations Simulator \cite{benninghoff2017european}. The main aim of the EPOS is to execute the cooperative rendezvous and docking system for Automated Transfer Vehicle (ATV) to the ISS. The EPOS test facilities consist of the test bed of two KUKA robots, 25m linear rail, sun simulation, realistic background, and HiLS setup for the execution of the end-to-end testing of rendezvous and docking. 

A similar concept has been used to develop our robotic space lab in TCS Research. The SRL is the first space robotics test facility in India that will be diversely working in the field of space robots. The test setup will test and validate cooperative and non-cooperative rendezvous and docking operations used in On-Orbit Servicing applications like Refueling, debris removal, maintenance, berthing, and assembly. The schematic diagram of the SRL lab is shown in figure \ref{space robotic lab layout}.

\begin{Figure}
    \centering
     \includegraphics[width=0.99\linewidth ]{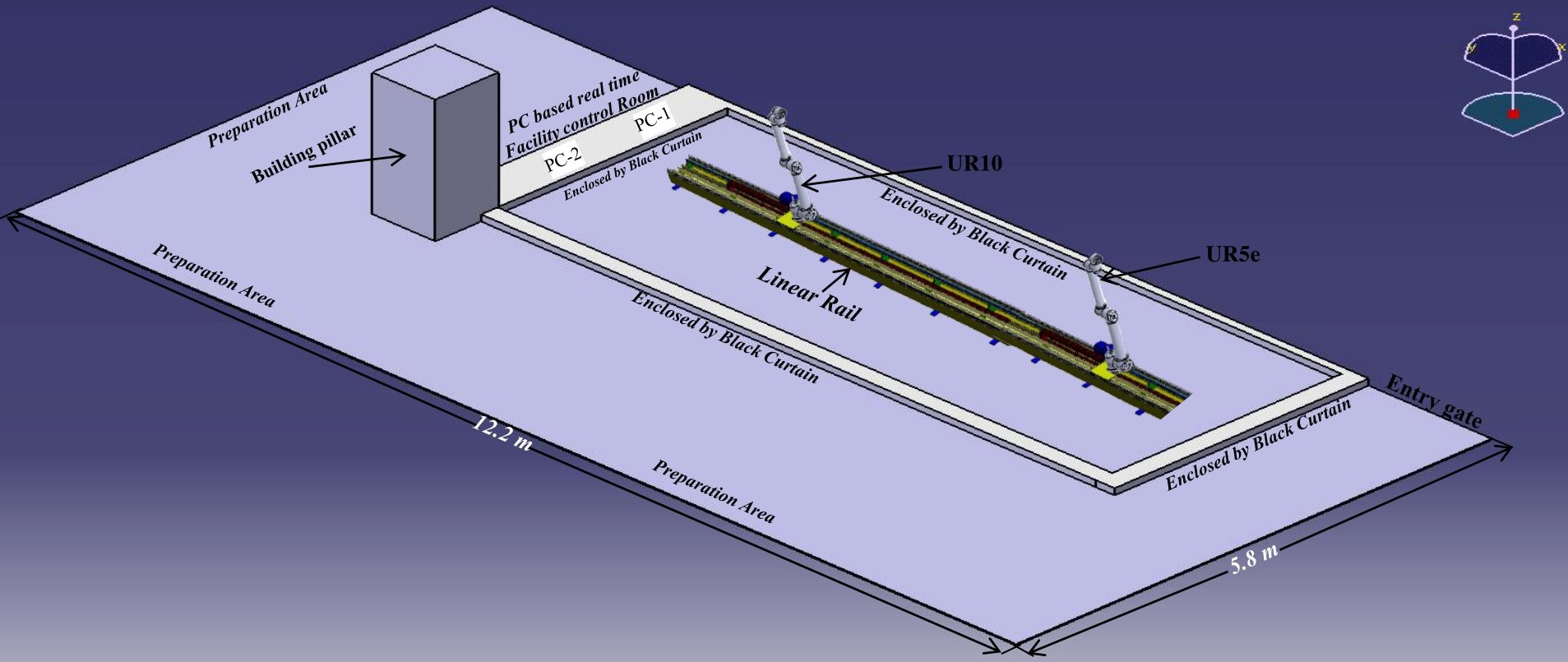}
      \captionof{figure}{ Space Robotics Lab Layout.  }
    \label{space robotic lab layout}
\end{Figure}

\subsection*{Objective}

The main objective of the SRL system is to test and verify the On-Orbit Servicing (OOS) tasks like refueling, assembly, debris removal, and maintenance by using the space robots. The experiment system typically considers the cooperative target, like a communication satellite that needs refueled or battery replacement or maintenance, and the non-cooperative target, like the spent satellite at LEO. The primary operations are near-field rendezvous, capture, fine manipulation, and docking to the target spacecraft for OOS.

The main experiment which will be carried out in the robotic space lab is listed below:

1. Testing and Verification of trajectory plan and control algorithm in service-target space robots.

2. Calibration, Testing, and Validation of force torque sensor in servicer space robot.

3. Testing and Verification of guidance algorithm while executing the proximity rendezvous between space robots

4. Testing and Validation of visual servoing algorithm to track non-cooperative targets.

5. Near field rendezvous and robotic capturing of non-cooperative targets autonomously.

6. Maintenance and replacement operation of faulty spacecraft in orbit.

\subsection*{Components of SRL.}

The space robotics lab in TCS research consists of the following components for the experiment:

\subsubsection*{Universal Robot Manipulator}

The universal robot manipulators set up the robotics lab's capture and fine manipulation operation. We will be using UR5e and UR10 manipulators for our experiment; the UR5e will act as the chaser space robot and will be replicating the chaser motion, and the UR10 will act as the target space robot, which will have the 3U mockup satellite model and will be replicating target motion. The lab UR5e and UR10 manipulators are shown in Figures \ref{fig:4} and \ref{fig:5}.

\begin{Figure}
    \centering
   \includegraphics[width = 0.99\linewidth]{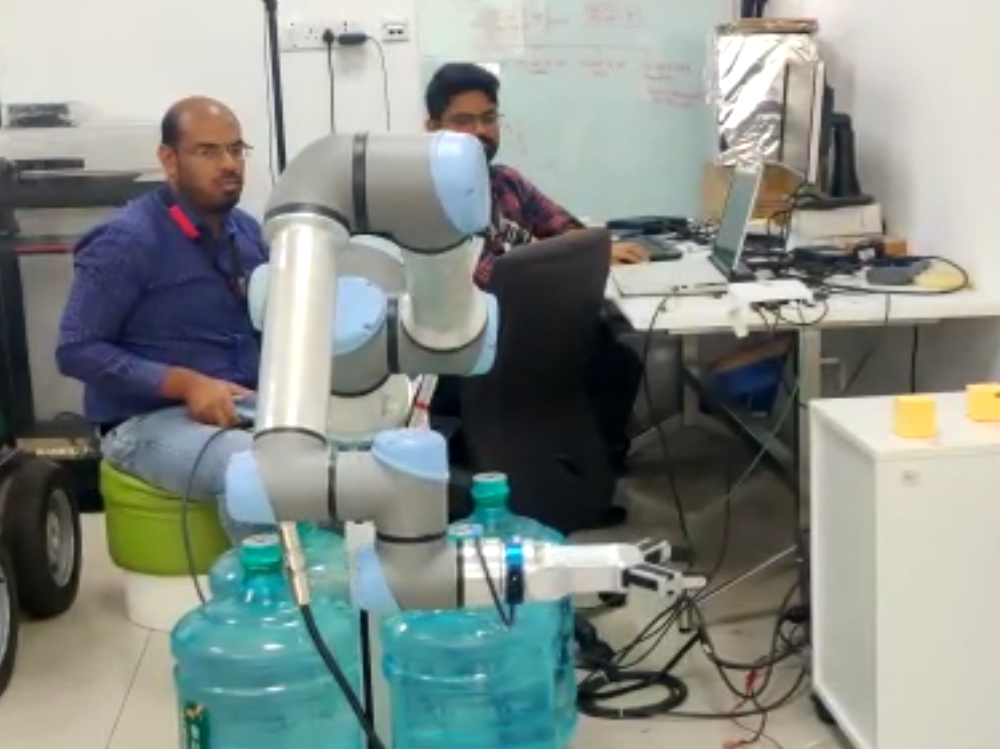}
    \captionof{figure}{UR5 manipulator replicating chaser motion.}
    \label{fig:4}
\end{Figure}

\begin{Figure}
    \centering
   \includegraphics[width = 0.99\linewidth]{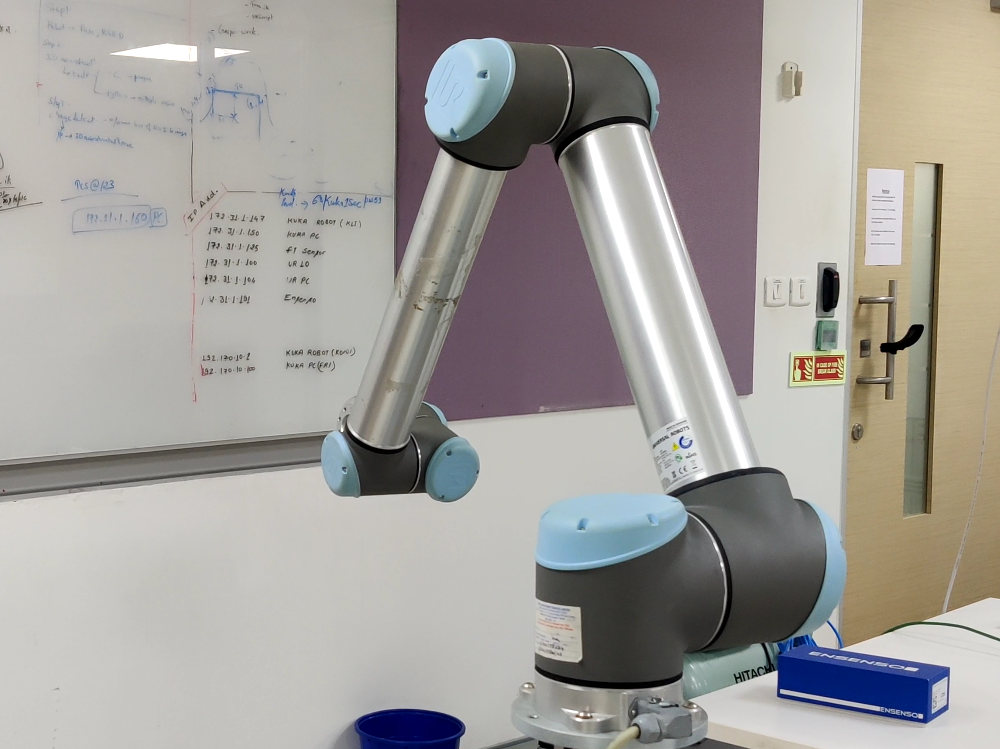}
    \captionof{figure}{UR10 manipulator replicating target motion.}
    \label{fig:5}
\end{Figure}

\subsubsection*{Linear Rail System } 

The 7th-axis linear rail system of 6 m in length is used to simulate the near-end rendezvous and proximity operation. A Rack and Pinion mechanism uses the servo-based control to move the URs robot in a linear direction. This linear rail system can withstand a payload of 20-30 kg with a speed of 1m/s and $\pm $ 0.1mm of positional accuracy. 

\begin{Figure}
    \centering
   \includegraphics[width = 0.99\linewidth]{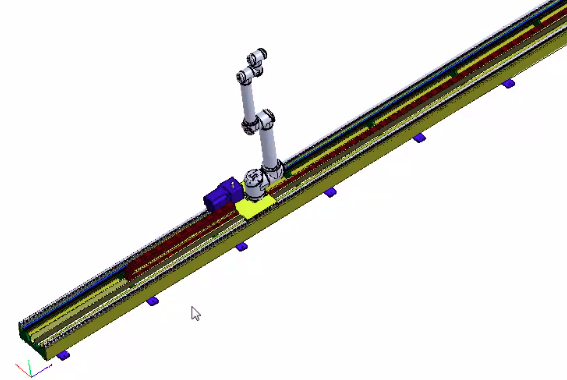}
    \captionof{figure}{Linear Rail System.}
    \label{linear rail}
\end{Figure}

The schematic diagram of the linear rail system is shown in figure \ref{linear rail}. The linear system is made of the mild-steel attached to the floor using the grouting method. In the linear rail system, the UR5e and UR10 supporting plates are there, which will help to provide linear movement to the robot. The relative velocity between the chaser and target object can be simulated using the linear rail system on a ground-based platform.

\subsubsection*{ Robotics Arm Gripper}

The robotics gripper aims to grasp the target object at its grasping points. An Onrobot RG2 gripper is used in chaser space robots which can withstand a payload of  2kg with a stroke of 110mm. The figure of the Onrobot RG2 gripper is shown in figure \ref{gripper} on the left side. This gripper has inbuilt firmware, which will be directly used by the URs manipulator end-effector and provides intelligence, fast deployment, easy customization, and programming.

\begin{Figure}
    \centering
    {{\includegraphics[trim=0cm 0cm 0cm 0cm,width=0.43\linewidth]{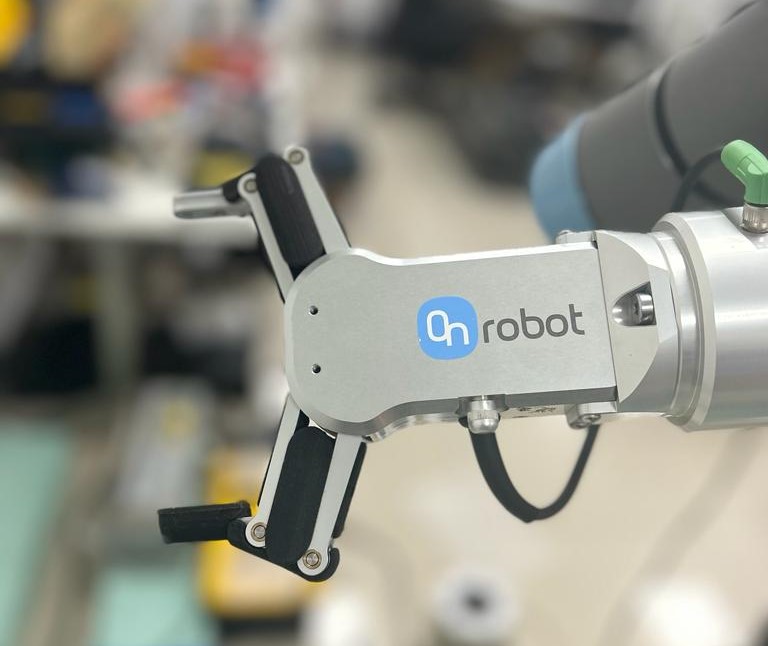}}}%
    \qquad
    {{\includegraphics[trim=0cm 0cm 0cm 0cm,width=0.43\linewidth, height =2.85 cm]{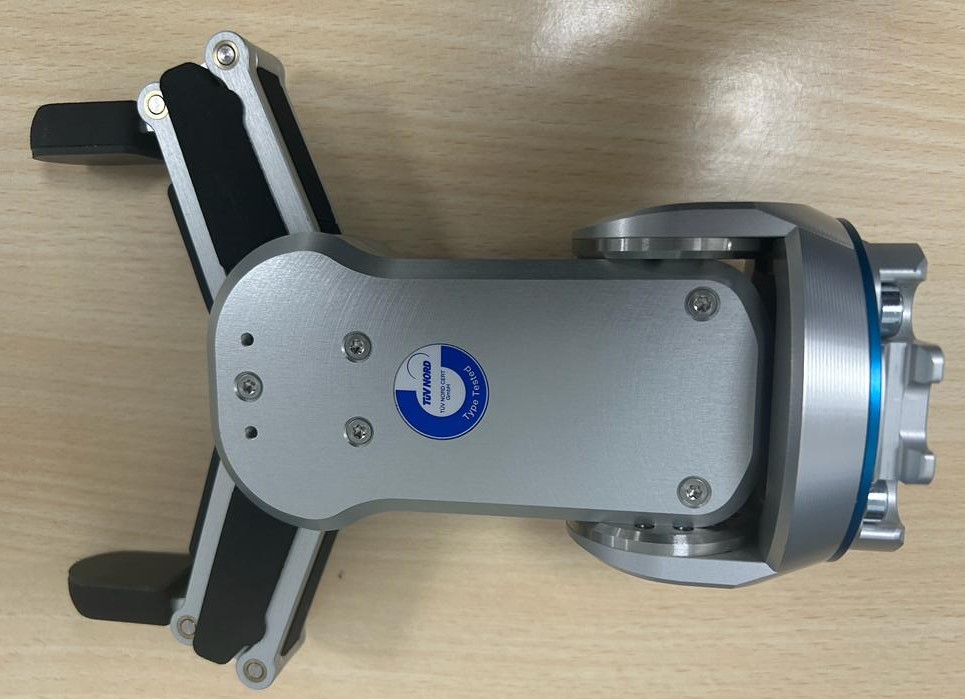}}}%
    \captionof{figure}{Robotic Grippers for UR5e and UR10.}
    \label{gripper}
\end{Figure}

Similarly, an Onrobot RG2 v2 gripper is used in the UR10 robot, where the mock-up satellite will be hung. This gripper, through UR10 control, primarily provides the tumbling motion to the 3U mock-up satellite body. The image of the Onboard RG2 v2 gripper is shown on the right side of figure \ref{gripper}.

\subsubsection*{Force Torque Sensor}

A Hex-E Onrobot 6-axis force/torque sensor is used for our UR5e manipulator. It provides accurate force and torque measurement along all 6DOF and gives us precise control for assembly, debris capturing, and refueling application. This sensor includes force control, path recording, high accuracy, and special features for insertion tasks. 

The Hex-E's primary function in our experiment is determining the impact force or torque acting on the chaser space robot end-effector while grasping the target object. It will help us to find the optimal impact force or torque acting on the chaser robot while docking and helps us to model the contact dynamics between the two space robot. This sensor can withstand the force of 220N and torques of 10Nm during the experimental phase. Figure \ref{Force torque sensor} shows the Hex-E force-torque sensor.

\begin{Figure}
    \centering
   \includegraphics[width = 0.99\linewidth]{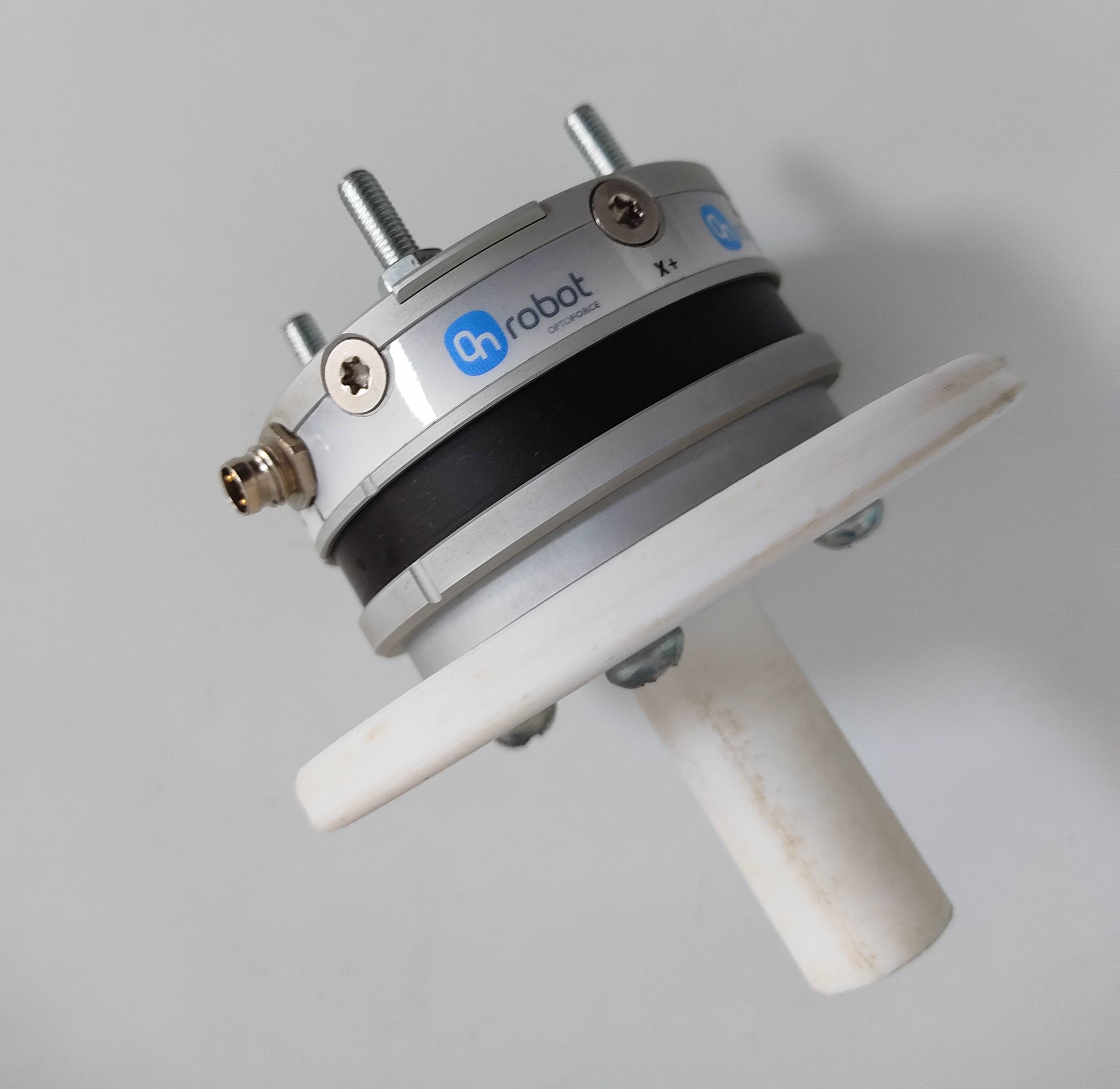}
    \captionof{figure}{Hex-E Onrobot 6-axis force/torque sensor.}
    \label{Force torque sensor}
\end{Figure}

\subsubsection*{3U Mock-up satellite}

A 3U standard CubeSat platform is used to make the mock-up satellite model. An acrylic sheet is used to make the 3U satellite model, and a solar panel is attached to the topmost surface of the model. Some grasping points have been made to the mock-up satellite model so that the gripper will grasp at given grasping points. The 3U satellite model is shown in figure \ref{3U mock-up}. The mock-up model will be placed on the UR10 manipulator, which provides the tumbling motion.

\begin{Figure}
    \centering
   \includegraphics[width = 0.6\linewidth, height = 3.0 in]{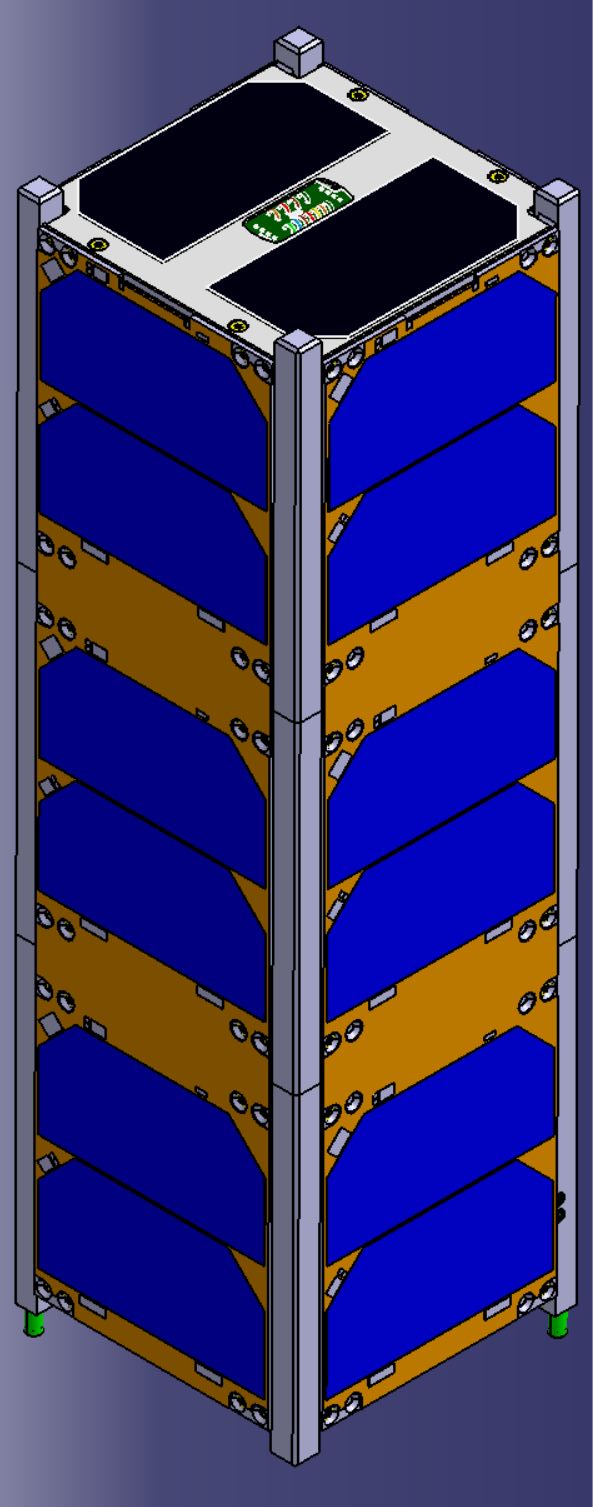}
    \captionof{figure}{3U mock-up satellite model.}
    \label{3U mock-up}
\end{Figure}

\subsubsection*{Vision Based Camera}

An Intel depth camera real sense D415 is used for the vision-based guidance and pose estimation of the target body from the chaser manipulator motion. Figure \ref{real sense camera} shows the depth camera which will be used for our HiLS experiment. This camera will be placed on the end-effector of the manipulator of the chaser. Its maximum range is approximately 10m, providing 1280 x 720 active stereo depth resolution with RGB image.

\begin{Figure}
    \centering
   \includegraphics[width = 0.99\linewidth]{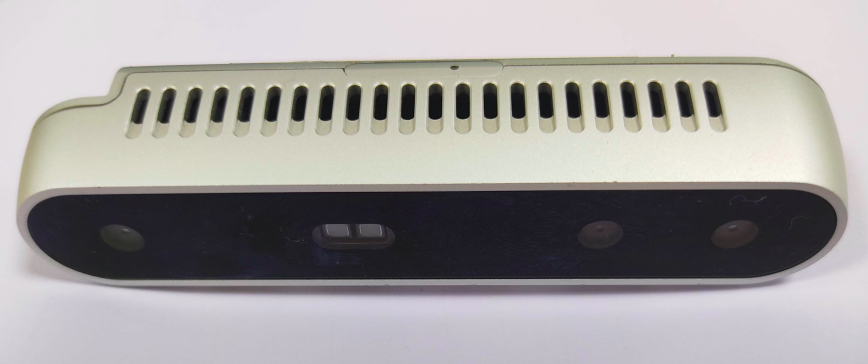}
    \captionof{figure}{Real Sense Depth Camera D415.}
    \label{real sense camera}
\end{Figure}

\begin{Figure}
    \centering
   \includegraphics[width = 0.99\linewidth]{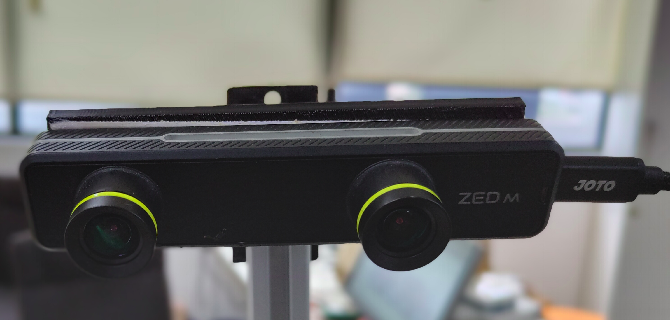}
    \captionof{figure}{Zed Camera.}
    \label{zed camera}
\end{Figure}

 An additional ZED camera is used for the motion sensing of the mock-up satellite model, which gives us 6DOF IMU for accurate motion tracking. The Zed camera used for our experiment is shown in figure \ref{zed camera}, and it provides ultra-depth sensing mode with high-speed data transfer and improved reliability. It has a small form factor of 6.5 cm eye separation, and provides an accurate 6DOF of a tumbling mock-up model.

\subsubsection*{Control Facility Room}

The space robotics lab consists of a PC-based real-time facility control room where 2 PCs will be used to inspect the UR5e and UR10 robots and control their motion in a linear rail platform. Tables and chairs will organize the control facility room with LED light movement, which will help us detect the target from different angles.

\subsubsection*{Miscellaneous accessories}

The miscellaneous accessories consist of a black curtain, LED light, preparation area,  window arrangement in the control facility room, wiring of 230V ac power, and GPU setup for running the hybrid experiment approach.
A black curtain encloses the experiment areas to mimic the space environment. An LED light setup is made at the experiment area, where the light will illuminate the target mock-up satellite to replica the Sunlight. The preparation area is mainly used for making plans and preparing the experimental procedures before performing any specified experiment.

\section*{RESULTS $\&$ DISCUSSION}
\label{Results}

The results section presents the complete pipeline of the workflow from the moment of the rendezvous phase of the rotation floating space robot ends. Once the target is in the workspace of the stowed manipulator arm, the Model Predictive Controller extends the arm with the gripper towards the target in the desired orientation. When this proximity operation ends, an on-off controller closes the gripper, followed by a PD controller, which brings the arm and the target to their initial orientation.

\subsection*{Preliminary Hardware-in-the-loop Results}

The hardware in the loop experiments is needed to validate the simulation results obtained from the bullet pipeline. Following the hybrid method philosophy described above, hardware experiments are independently conducted to replicate the chaser arm motion on a UR5 arm and target motion on a UR 10 arm. As explained, these motions concern a frame attached to the base satellite in the PyBullet environment. The feedback is obtained from the arms and given to the PyBullet to get the MPC command and update the system's dynamics. Figure \ref{UR5_motion} shows the motion of the UR5 arm while replicating the chaser motion leading to the plot \ref{Hardware_Chaser} obtained. In figure \ref{UR5_motion} a.,  the initial hardware motion of the UR5 manipulator is shown, which shows the deviation from the Pybullet simulation command. As the chaser motion proceeds, the UR5 starts following the trajectory similar to the Pybullet simulation, shown in figure \ref{UR5_motion}b. and figure \ref{UR5_motion}c. Although stage II and stage III show the difference in x,y, and z position values with the simulation results, the trajectory path remains the same.

\begin{Figure}
    \centering
   
    {{\includegraphics[trim=0cm 0cm 0cm 0cm,width=0.96\linewidth]{results/UR5_1.png}}}
    \captionof*{figure}{a. Stage I}
    \centering
\end{Figure}

\begin{Figure}
    \centering

    {{\includegraphics[trim=0cm 0cm 0cm 0cm,width=0.96\linewidth]{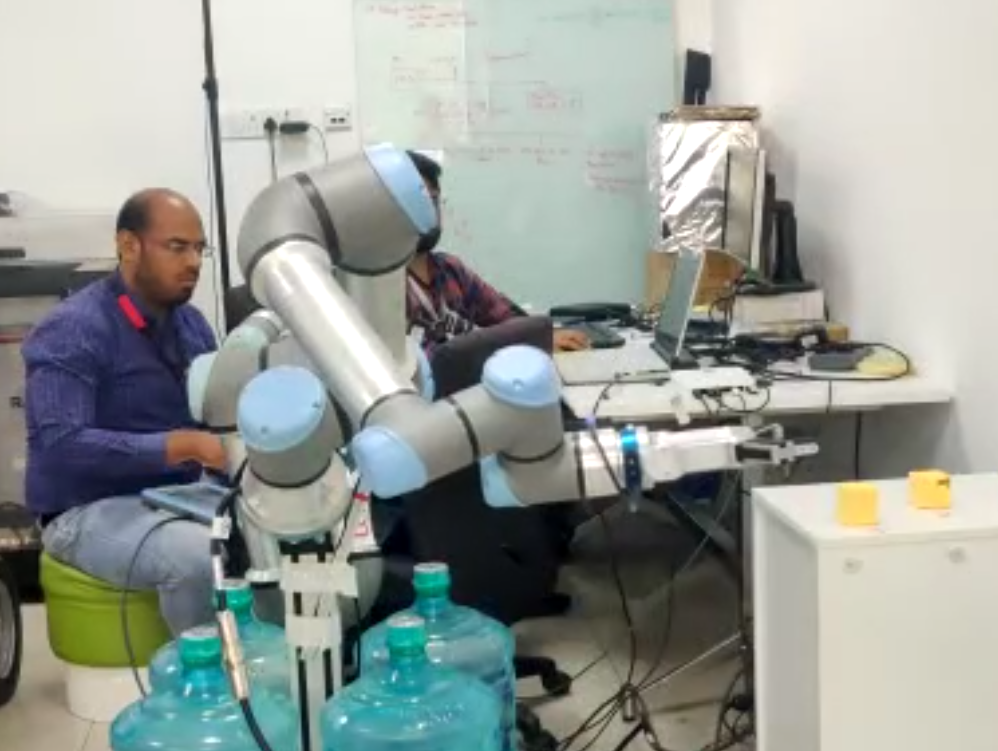}}}
    \captionof*{figure}{b. Stage II}
    \centering
\end{Figure}

\begin{Figure}
    \centering

    {{\includegraphics[trim=0cm 0cm 0cm 0cm,width=0.96\linewidth]{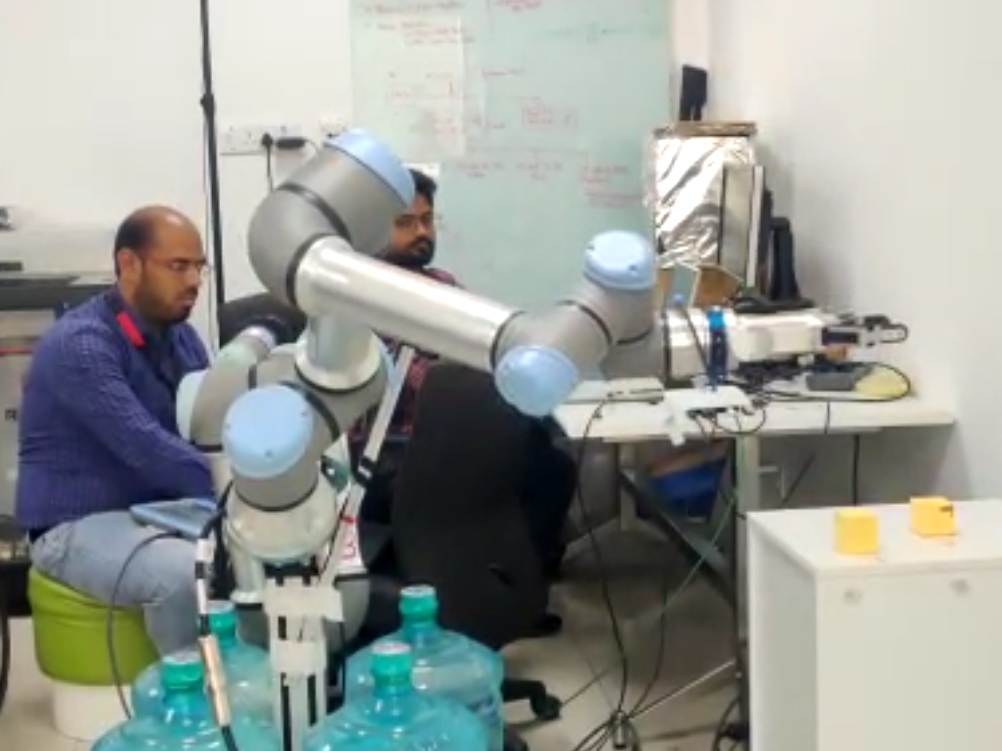}}}
    \captionof*{figure}{c. Stage III}
    \centering
\end{Figure}

Figure \ref{UR5_motion} d. shows the final position of the UR5, similar to the Pybullet simulation results. It was found that the UR5 is reaching the target point similar to the Pybullet trajectory path with slight deviation caused by the time latency, hardware friction, etc.

\begin{Figure}
    \centering
    {{\includegraphics[trim=0cm 0cm 0cm 0cm,width=0.96\linewidth]{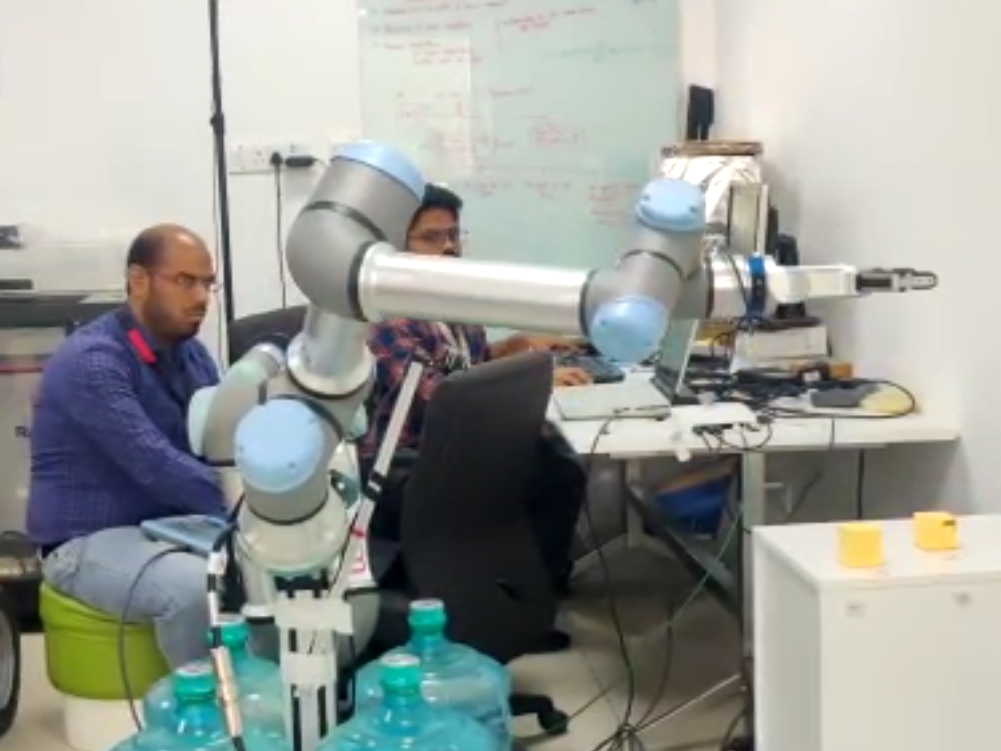}}}
    \captionof*{figure}{d. Stage IV}
    \centering
     \captionof{figure}{UR5 motion while replicating chaser motion.}
    \label{UR5_motion}
\end{Figure}

 Figure \ref{Hardware_Chaser} shows the motion of the UR5 arm replicating the chaser arm motion. Solid lines are the actual trajectories of the UR5 arm, while dotted lines are the trajectories computed by the MPC, which are given as commands to the UR5 robot. It can be seen that the chaser can follow the commanded trajectory with good precision. 

\begin{Figure}
    \centering
    {{\includegraphics[trim=0cm 0cm 0cm 0cm,width=0.99\linewidth]{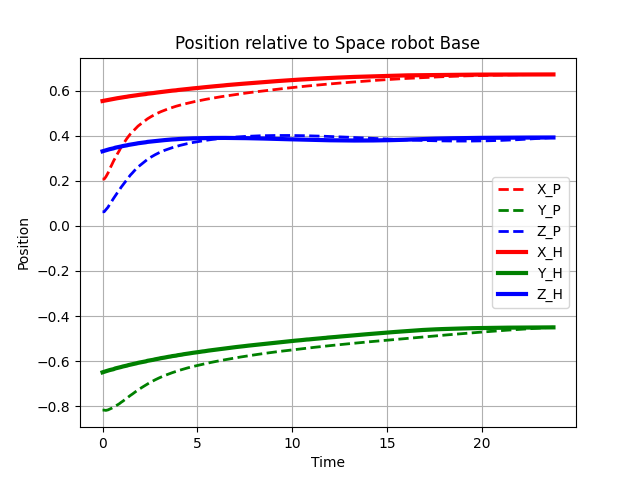}}}
    \captionof{figure}{Commanded vs Followed Position of UR5 arm representing Chaser.}
    \label{Hardware_Chaser}
\end{Figure}

Figure \ref{UR10_motion} shows the motion of the UR10 arm while replicating the chaser motion leading to the plot \ref{Hardware_Target} obtained. In figure \ref{UR10_motion} a., the initial hardware motion of the UR5 manipulator is shown, which shows the deviation from the Pybullet simulation command. As the target motion proceeds, the UR10 follows the trajectory similar to the Pybullet simulation, shown in figure \ref{UR10_motion} b. and figure \ref{UR10_motion} c. Although stage II and stage III show the difference in x,y, and z position values with the simulation results, the trajectory path remains the same.

\begin{Figure}
    \centering

    {{\includegraphics[trim=0cm 0cm 0cm 0cm,width=0.96\linewidth]{results/UR10_1.png}}}
    \captionof*{figure}{a. Stage I}
    \centering
\end{Figure}

\begin{Figure}
    \centering

    {{\includegraphics[trim=0cm 0cm 0cm 0cm,width=0.96\linewidth]{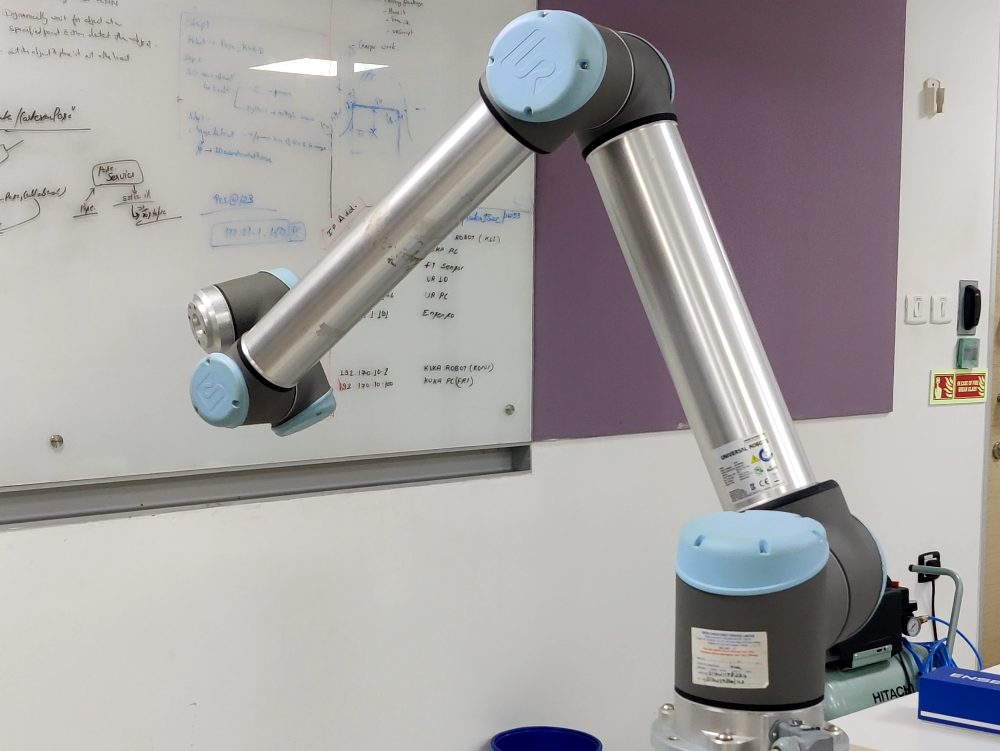}}}
    \captionof*{figure}{b. Stage II}
    \centering
\end{Figure}

\begin{Figure}
    \centering

    {{\includegraphics[trim=0cm 0cm 0cm 0cm,width=0.96\linewidth]{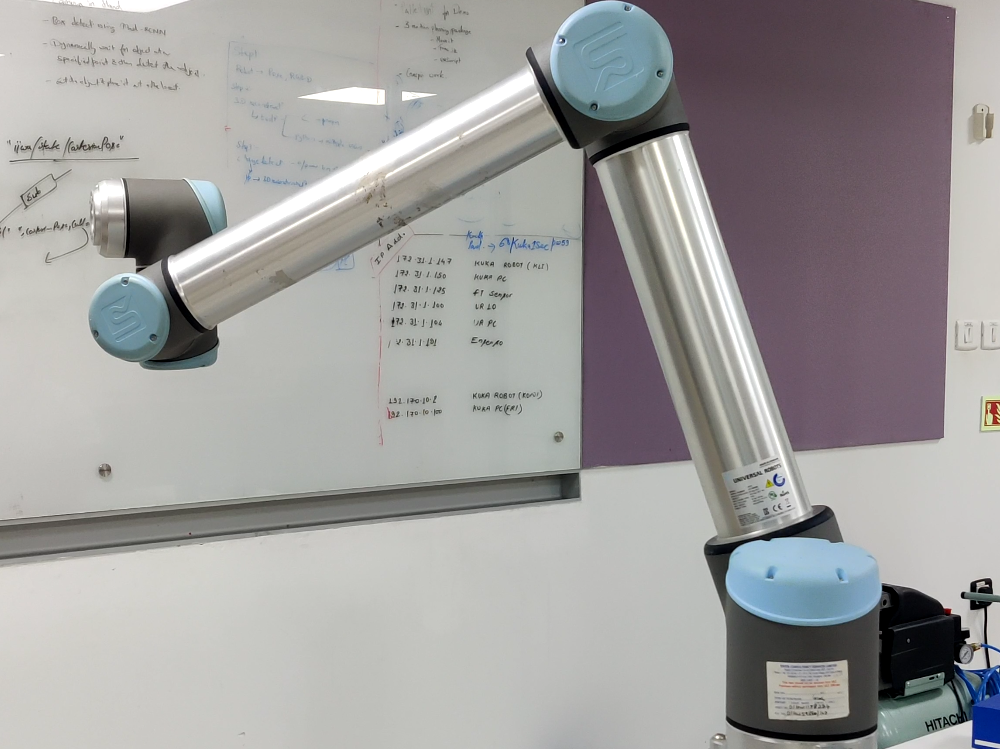}}}
    \captionof*{figure}{c. Stage III}
    \centering
\end{Figure}

Figure \ref{UR10_motion} d.  shows the final position of the UR5, similar to the Pybullet simulation results. It was found that the UR10 is reaching the target point similar to the Pybullet trajectory path with slight deviation caused by the time latency, hardware friction, etc.
\begin{Figure}
    \centering
    {{\includegraphics[trim=0cm 0cm 0cm 0cm,width=0.96\linewidth]{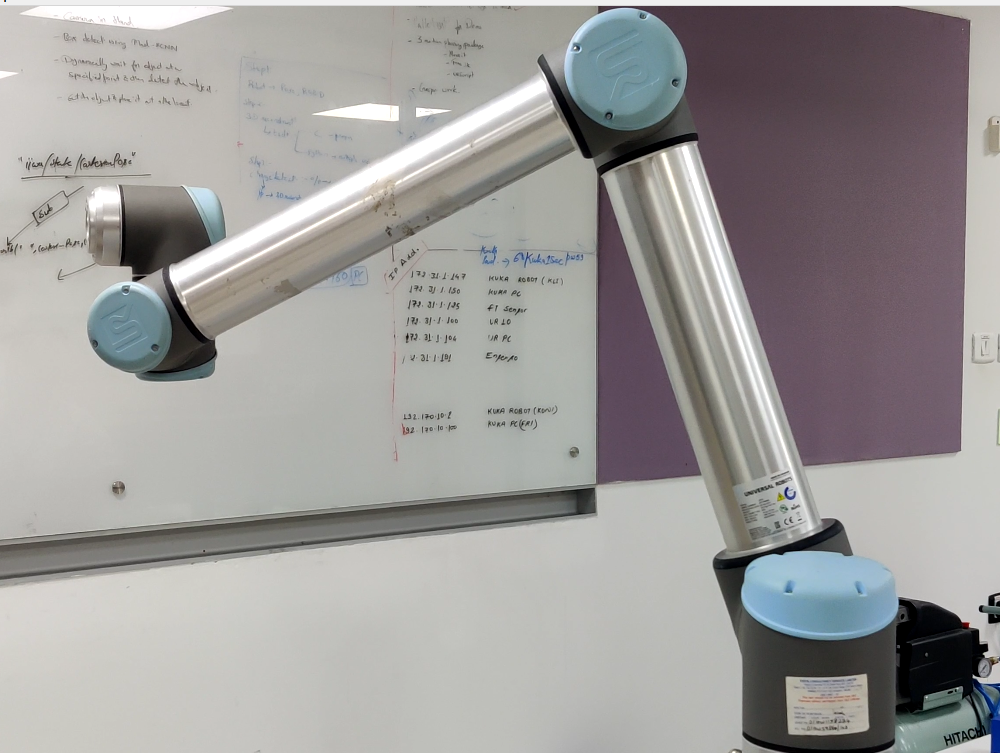}}}
    \captionof*{figure}{d. Stage IV}
    \centering
     \captionof{figure}{UR10 motion while replicating target motion.}
    \label{UR10_motion}
\end{Figure}

Similar results are obtained for the target motion using the UR10 arm. Figure \ref{Hardware_Target} shows the motion of the UR10 arm (solid lines) against the motion of the target obtained from PyBullet. These motions concern an intermediate frame attached to the base satellite in PyBullet.

\begin{Figure}
    \centering
    {{\includegraphics[trim=0cm 0cm 0cm 0cm,width=0.99\linewidth]{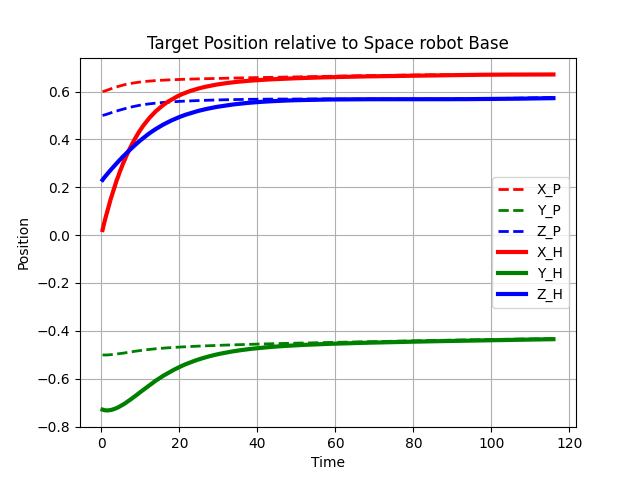}}}
    \captionof{figure}{Commanded vs Followed Position of UR10 arm representing Target.}
    \label{Hardware_Target}
\end{Figure}

We intend to complete the PyBullet pipeline on the hardware-in-the-loop setup for the final results. For the same, we are developing a mock satellite model to act as the target. Force and torque sensors will be applied on the walls of the UR5 (chaser gripper), and the values will be recorded. These will help us design an impedance controller for the contact dynamics.

\section*{CONCLUSION $\&$ FUTURE WORK}

The paper presents the development process of the HiLS facility setup for the space robotics application. TCS Research, India's first space robotics laboratory, has developed the space robotics lab to execute the proximity rendezvous and docking process. Numerous objective has been set up that will be implemented in the space robotics lab. A hybrid simulation approach has been utilized to emulate the micro-gravity environment on the ground with minimum cost. This approach has been promising because free base motion will be captured in a simulation environment like Pybullet, and the manipulators will mimic the chaser's and target's motion in hardware. UR5e and UR10 manipulators are used to replicate the motion of the chaser and target object. A 3U CubeSat mock-up model is made, which will be mounted on UR10 and provide the tumbling motion to it. The MPC controller extends the arm with the gripper as an end effector toward the target in the desired orientation. Once reaching the proximity,  an on-off controller closes the gripper, followed by a PD controller, which brings the arm and the target to their initial orientation. The paper also presents the initial results of replicating the chaser and target motion by UR5 and UR10 manipulators. The results show that the motion of the UR5 and UR10 arms against the motion of the chaser and target obtained from the Pybullets are in the same trajectory line. 

In future work, we intend to develop the mock-up model and mount it at UR10, which will act as a target. A plan to add the force torque sensor will be applied on the end-effector of UR5, i.e., the chaser manipulator, and the values will be recorded while capturing the target objects. The multiple scenarios will be run using an F/T sensor to find efficient values to design the impedance controller for the contact dynamics. In the later phase of the experiment, a confirmed case of the thrust characteristics \cite{AIP2020} \cite{Ros_2021} \cite{kumarinvestigation} \cite{sah2017design} obtained from the STK simulation  \cite{VETOMAC2022}   \cite{Roshan_2021_studsat} will be incorporated into the chaser manipulator to execute the rendezvous trajectory close to the target object by using the linear rail.

\bibliography{sample.bib}
\bibliographystyle{ieeetr}

\end{multicols*}
\end{document}